%% file: paper.tex
\documentclass[]{TEAI}
\usepackage{helvet}

\usepackage{amsmath} 
\usepackage{natbib}
\usepackage{graphicx}
\usepackage{subcaption} 

\usepackage[toc,page,header]{appendix}
\usepackage[utf8]{inputenc} % allow utf-8 input
\usepackage[T1]{fontenc}    % use 8-bit T1 fonts
\usepackage{hyperref}       % hyperlinks
\usepackage{url}            % simple URL typesetting
\usepackage{booktabs}       % professional-quality tables
\usepackage{amsfonts}       % blackboard math symbols
\usepackage{nicefrac}       % compact symbols for 1/2, etc.
\usepackage{microtype}      % microtypography
\usepackage{wrapfig}

\usepackage{amssymb}  % 用于特殊符号如♠♣等
\usepackage{fontawesome}  % 用于邮箱图标 \faEnvelope
\usepackage{url}  % 用于 \url 命令

\usepackage{titletoc}

\usepackage{tikz}  % 用于绘制彩色标记符号
\usepackage{comment}  % 用于注释备用版本
\usepackage{tabularx}  % 如果需要自动调整列宽
\usepackage{booktabs}  % 用于更美观的表格线条(可选)
\usepackage{minitoc}

\usepackage{booktabs}
\usepackage{array}
\usepackage{etoolbox}

\definecolor{lightblue}{RGB}{200, 230, 255}  
\definecolor{headerblue}{RGB}{150, 200, 255} 

\usepackage{pgfplots}
\pgfplotsset{compat=1.18}
\usepackage[utf8]{inputenc} % allow utf-8 input
\usepackage[T1]{fontenc}    % use 8-bit T1 fonts
\usepackage{hyperref}       % hyperlinks
\usepackage{url}            % simple URL typesetting
\usepackage{booktabs}       % professional-quality tables
\usepackage{amsfonts}       % blackboard math symbols
\usepackage{nicefrac}       % compact symbols for 1/2, etc.
\usepackage{microtype}      % microtypography
\usepackage{xcolor}         % colors
\usepackage{graphicx}
\usepackage{float}
\usepackage{comment}
\usepackage{multirow} % For multi-row cells
\usepackage{amsmath} % For \text command if needed inside math mode\Delta
\usepackage{makecell} % For multi-line cells and better vertical spacing in cells
\usepackage{siunitx}  % For better number alignment (optional but recommended)
\usepackage{tikz}
\usepackage{pgf-pie} % Package for creating pie charts
\usepackage{subcaption}
\usepackage{wrapfig}
\usepackage[export]{adjustbox}

\usepackage{ragged2e}      % for \RaggedRight in tabularx
\usepackage{tabularx}       % For tables with fixed total width and auto-adjusting columns
\usepackage{array}          % For advanced column formatting (like >{\centering\arraybackslash}X)
\usepackage{caption}        % Recommended for figures/tables, but we'll do simple text below images here.
\usepackage{enumitem}
\usepackage{pifont}
\usepackage[hang,flushmargin]{footmisc} % 更好的脚注处理

\usepackage{tcolorbox}
\usepackage{tcolorbox}    % 1. 先加载 tcolorbox 主包
\tcbuselibrary{breakable}  % 2. 显式加载 breakable 库
\tcbuselibrary{skins}      % 3. 显式加载 skins 库 (这个库提供 topruleatbreak 等选项)
\usepackage{tabularx}
\usepackage{listings}

\usepackage{algorithm}
\usepackage{algpseudocode}

\tcbuselibrary{listings}

\newtcblisting{promptbox}[1]{
    enhanced,
    breakable,
    listing only,
    title={#1},
    colback=gray!5,
    colframe=gray!60,
    boxrule=0.5pt,
    arc=2pt,
    listing options={
        basicstyle=\ttfamily\small,
        breaklines=true,
        columns=fullflexible,
        keepspaces=true,
        showstringspaces=false
    }
}

\title{Spatial-Zero: Self-Evolving Spatial Reasoning 
in Vision-Language Models from Zero Data}

\author{
    Yutong Li\textsuperscript{1},
    Haoyu Zhao\textsuperscript{2},
    Daoguo Dong\textsuperscript{1},
    Ziyi Ye\textsuperscript{1},
    Yu-gang Jiang\textsuperscript{1}
}

\affiliation[1]{Institute of Trustworthy Embodied AI, Fudan University}
\affiliation[2]{National University of Singapore}

\abstract{\input{sec/0_Abstract}}

\checkdata[Code]{\url{https://github.com/YutongLi2024/Spatial-Zero}}
\checkdata[Contact Email]{\email{liyt26@m.fudan.edu.cn}}

\newcommand{\mainsec}[1]{%
    \noindent\textit{Main paper: Sec.~#1.}\par
}
\newcommand{\Qvalid}{\mathcal{Q}_{\mathrm{valid}}}
\newcommand{\GT}{y^{\mathrm{GT}}}
\newcommand{\ind}{\mathbb{I}}

\begin{document}
\maketitle

\vspace{-1.5em}
\input{sec/1_Introduction}
\input{sec/3_Methodology}

\input{sec/4_Experiment}
\input{sec/2_RelatedWork}

\input{sec/5_Conclusion}

\bibliographystyle{plainnat}
\bibliography{main}

\input{sec/X_Appendix}

% \clearpage

\end{document}

%% file: sec/0_Abstract.tex
\begin{abstract}
SSpatial reasoning requires a vision-language model~(VLM) to understand metric distances, three-dimensional relationships, and viewpoint-dependent geometry. 
Learning these capabilities typically requires externally curated training data.
This paper introduces Spatial-Zero, a VLMs' training framework that can self-evolve through autonomous spatial exploration with zero data. 
Spatial-Zero addresses two key challenges in autonomous spatial exploration: reliably generating valid spatial interaction tasks and providing accurate supervision for learning from these tasks.
Specifically, a proposer specifies object configurations and questions, and a Coder turns them into executable scene programs that produce multi-view images and explicit scene geometry. 
A Geometric Oracle checks spatial constraints, visibility, and answer uniqueness and derives ground-truth labels from executed geometry; these labels supervise the Solver and ground difficulty feedback, while validation outcomes guide subsequent task generation. 
Across six spatial reasoning benchmarks and three VLM backbones, Spatial-Zero yields consistent gains through iterative self-evolution. After three iterations, Qwen3.5-9B improves average accuracy by 3.23 percentage points; with Qwen3.5-4B, Spatial-Zero also outperforms existing VLM self-evolution methods.
% Our code is available at~\url{https://github.com/YutongLi2024/Spatial-Zero}
\end{abstract}

%% file: sec/1_Introduction.tex
\begin{figure*}[ht]
    \centering
    \includegraphics[width=\linewidth]{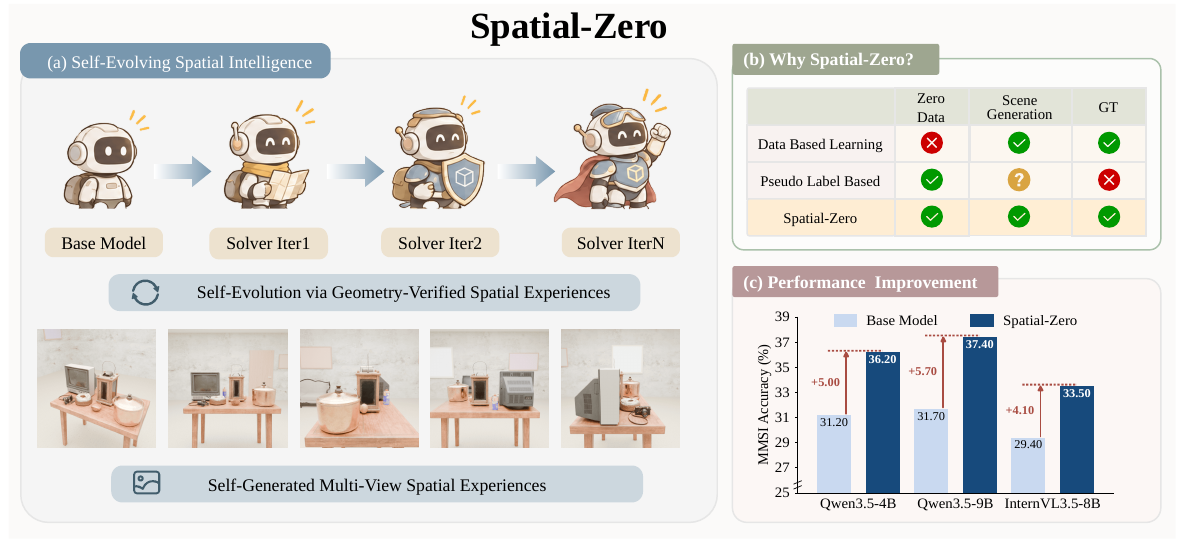}
    \caption{Overview of Spatial-Zero.
    (a) Self-generated, geometry-verified spatial experiences improve spatial reasoning.
    (b) Comparison of learning paradigms in data dependence, scene generation, and ground-truth supervision.
    (c) MMSI improvements across three backbones.}
    \label{fig:introduction}
\end{figure*}

\section{Introduction}

Spatial reasoning is fundamental to understanding and interacting with the physical world, underpinning applications such as embodied navigation, robotic manipulation, and autonomous driving~\cite{team2026qwen3, Robospatial, Spacetools, SpatialVLM}. Despite advances in vision-language models (VLMs)~\cite{yang2025qwen3, qwen3.5, LLaVA-OneVision, team2026kimi}, interpreting metric distances, three-dimensional object relationships, and viewpoint-dependent geometry remains challenging~\cite{Viewspatial-bench, embspatial-bench, Robospatial, Cambrian-s}. These capabilities require reasoning about physical scene structure beyond visual appearance.

Motivated by the role of environmental interaction in human spatial learning, recent studies have explored self-evolution to improve both general visual reasoning~\cite{V-Zero,VisPlay,MM-Zero_EMNLP2026} and spatial reasoning in VLMs~\cite{SMA,spatialskill,Spatialevo,SAGE}. These approaches iteratively generate learning tasks and feedback, yet face two recurring dependencies, as illustrated in Fig.~\ref{fig:introduction}(b). First, data-based learning relies on externally provided images, spatial observations, or reconstructed 3D scenes to construct training experiences~\cite{V-Zero,VisPlay,Spatialevo}. Second, pseudo-label-based learning derives supervision from model predictions, often through majority voting over sampled responses rather than independently verified ground truth~\cite{V-Zero,MM-Zero_EMNLP2026}. For example, MM-Zero takes a step toward zero-data self-evolution by generating visual tasks through executable programs, yet still relies on model-voted labels for challenging questions~\cite{MM-Zero_EMNLP2026}.

However, extending zero-data visual self-evolution to 3D spatial reasoning requires more than replacing 2D visual synthesis with 3D rendering. First, visually plausible content does not guarantee valid spatial experiences, which require consistent 3D configurations, satisfied spatial constraints, sufficient visibility, and uniquely answerable questions~\cite{SpatialVLM, VSI-Bench, 3DSRBench}. Although MM-Zero employs rendering and simple-question verification, these mechanisms do not ensure geometric validity. Second, reliable spatial self-evolution requires correctness-grounded supervision rather than model consensus alone~\cite{Spatialevo}. MM-Zero uses majority-voted pseudo-labels for Solver training and prediction consistency to estimate question difficulty, potentially reinforcing systematic spatial errors~\cite{MM-Zero_EMNLP2026}. The importance of geometric ground truth is highlighted by SpatialEvo, where replacing geometry-derived labels with majority-voted pseudo-labels reduces VSI-Bench accuracy from 46.1 to 18.8~\cite{Spatialevo}. Nevertheless, SpatialEvo relies on pre-existing reconstructed scenes for geometric supervision, constraining spatial configurations to external resources. These limitations raise a question: how can we autonomously generate geometrically valid 3D spatial experiences, derive trustworthy ground truth, and continuously evolve these experiences through reliable feedback without external spatial training data?

To address this challenge, we introduce \textbf{Spatial-Zero}, a zero-data self-evolving framework that couples autonomous 3D scene generation with deterministic geometric supervision. Spatial-Zero iteratively optimizes three roles initialized from the same VLM backbone. As illustrated in Fig.~\ref{fig:introduction}(a), the Proposer jointly specifies spatial configurations and questions, while the Coder translates these specifications into executable 3D scene programs. Program execution produces multi-view images together with explicit object geometry and camera parameters. A deterministic Geometric Oracle validates the generated scenes and questions against geometric constraints, visibility, and answer uniqueness, and derives ground-truth labels directly from the executed geometry. Crucially, these labels not only supervise the Solver but also enable difficulty feedback grounded in actual correctness, while geometric validation and execution feedback guide the subsequent generation of realizable and diverse spatial experiences. Through iterative optimization, the Proposer and Coder refine spatial configurations and questions in response to validation outcomes and Solver performance, while the Solver improves through geometry-verified supervision. As highlighted in Fig.~\ref{fig:introduction}(b), Spatial-Zero establishes a closed learning loop in which spatial experiences themselves evolve without relying on pre-existing training scenes or model-voted pseudo-labels.

Extensive experiments across six spatial reasoning benchmarks demonstrate that Spatial-Zero consistently improves the spatial reasoning capabilities of different VLM backbones through iterative self-evolution. For example, Qwen3.5-9B achieves a 3.23-point improvement in average accuracy after three iterations, while Spatial-Zero also outperforms existing VLM self-evolution methods under the same Qwen3.5-4B backbone. As illustrated in Fig.~\ref{fig:introduction}(c), consistent improvements on MMSI across three VLM backbones further demonstrate the effectiveness of geometry-verified self-evolution for spatial reasoning.

Our main contributions are summarized as follows:

\begin{itemize}
    \item We propose Spatial-Zero, a zero-data self-evolving framework that enables autonomous generation and iterative refinement of 3D spatial learning experiences.
    \item We integrate deterministic geometric verification into self-evolution, providing reliable ground-truth supervision and feedback for spatial experience generation without external spatial training data.
    \item Extensive experiments across six spatial reasoning benchmarks and three VLM backbones demonstrate consistent improvements, validating the effectiveness of geometric supervision and iterative self-evolution.
\end{itemize}

%% file: sec/3_Methodology.tex
\begin{figure*}[ht]
    \centering
    \includegraphics[width=1\linewidth]{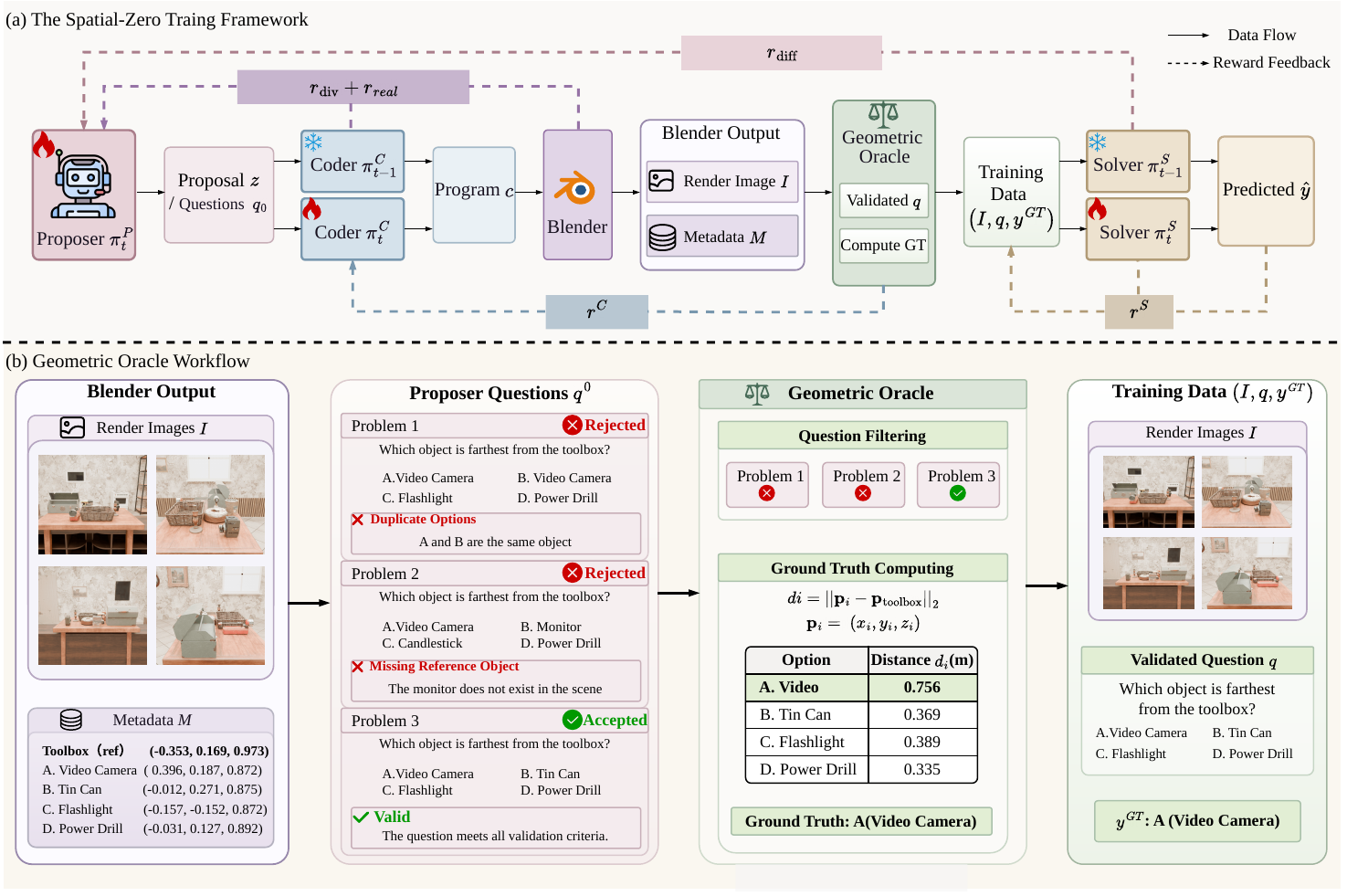}
    \caption{
    Overview of the Spatial-Zero framework. 
    (a) The model training framework.
    (b) Geometric Oracle workflow.
    }
    \label{fig:spatial_zero_framework}
\end{figure*}

\section{Methodology}\label{sec:methodology}

\subsection{Preliminaries}\label{sec:preliminaries}
\paragraph{Reinforcement Learning with Verifiable Rewards.} 
Reinforcement learning with verifiable rewards (RLVR) improves language model reasoning by optimizing policies with objectively verifiable feedback rather than learned reward models or human preference judgments~\cite{Deepseekmath, RLVR}. Given an input \(x\) and a model-generated response \(y\), a verifier assigns a reward \(r(x,y)\) according to predefined correctness criteria. This paradigm has demonstrated effectiveness in domains such as mathematical reasoning and code generation, where answers can be verified through exact matching or program execution.

\paragraph{Group Relative Policy Optimization.}
We optimize all three trainable roles with Group Relative Policy Optimization (GRPO)~\cite{Deepseekmath}. For a role $\rho\in\{P,C,S\}$, the rollout policy $\pi_{\mathrm{old}}^{\rho}$ generates a group of $G$ responses $\{y_i\}_{i=1}^G$ to an input $x$, with role-specific rewards $\{r_i^\rho\}_{i=1}^G$. Relative advantages are estimated as
\begin{equation}
    \hat A_i^\rho=\frac{r_i^\rho-\mu_x^\rho}{s_x^\rho+\epsilon},
    \qquad \mu_x^\rho=\frac{1}{G}\sum_{j=1}^{G}r_j^\rho,
    \label{eq:grpo_advantage}
\end{equation}
where $\epsilon$ is a numerical stabilizer. Following our optimization setting, $s_x^\rho$ is the group-level reward standard deviation for Proposer and Solver and the batch-level reward standard deviation for Coder. The resulting advantages are used in a clipped policy objective with KL regularization, without training a separate value model.

\subsection{Overview}\label{sec:overview}
Spatial-Zero improves 3D spatial reasoning without externally supplied spatial training examples. Figure~\ref{fig:spatial_zero_framework} illustrates the overall framework and training workflow of Spatial-Zero. Spatial-Zero initializes three distinct policies, Proposer, Coder, and Solver, from the same vision-language model. The Proposer generates scene configurations and spatial questions, while the Coder translates these proposals into executable Blender programs. The Geometric Oracle executes and validates the generated scenes, producing multi-view images and computing ground-truth answers from the realized geometry. The Solver learns to answer validated questions based on rendered images, with geometric ground truth providing verifiable rewards. The three policies are optimized sequentially using GRPO, forming a self-evolution process without reliance on external spatial annotations or model-generated pseudo-labels.

\subsection{Spatial-Zero Training Pipeline}\label{sec:training_pipeline}
Let $\pi_t^P$, $\pi_t^C$, and $\pi_t^S$ denote the three policies after iteration $t$, all initialized from the same base model at $t=0$. At each iteration $t\geq 1$, we update the policies sequentially in the order $P\rightarrow C\rightarrow S$ using GRPO. Each role is optimized from its preceding checkpoint, while other policies used to generate inputs or assess outputs remain frozen in that phase. The Geometric Oracle is deterministic and never updated.

\paragraph{Proposer.}
The Proposer generates proposals $z$ specifying indoor or outdoor environments, target objects, spatial relations, and a set of multiple-choice questions $q_0$ with options but no answer labels. The environment is selected by the Proposer rather than fixed per iteration. The frozen previous Coder $\pi_{t-1}^C$ attempts scene realization, while the Geometric Oracle validates questions and derives ground-truth answers from geometry. The frozen previous Solver $\pi_{t-1}^S$ is evaluated against these answers to estimate difficulty. Rewards encourage challenging, numerous and diverse valid questions, and realizable scenes. Invalid proposals receive $-1$, and invalid questions are not repaired. Further details on proposal representation and validation are provided in Appendix~\ref{app:proposal}.

\paragraph{Coder.}
The updated Proposer $\pi_t^P$ provides valid, deduplicated proposals for Coder training. Conditioned on each proposal, the Coder generates Blender programs specifying object placement and attributes. Programs are evaluated for executability, legal placement, and spatial constraint satisfaction. Violations directly reduce the Coder reward without modifying the generated scene, while unusable programs incur a failure penalty. Only candidates passing the required checks are retained for rendering. Further details on Coder scene construction and execution are provided in Appendix~\ref{app:coder}.

\paragraph{Geometric Oracle.}
The Geometric Oracle operates on a scene program $c$ and its associated proposal $z$. It executes $c$ in Blender to obtain multi-view observations $\mathcal I$ and scene metadata $M$ containing object geometry, camera parameters, and visibility information. Based on the executed geometry, the oracle verifies scene constraints and validates the proposed question set $q_0$, retaining valid questions $q\in\mathcal Q_{\mathrm{valid}}\subseteq q_0$. For each validated question, the oracle computes the ground-truth answer:
\begin{equation}
    y^{\mathrm{GT}}=\Omega(M,q),
    \qquad (\mathcal I,q,y^{\mathrm{GT}}).
    \label{eq:geometry_oracle}
\end{equation}
Here, $y^{\mathrm{GT}}$ is obtained by evaluating spatial relations in the executed 3D scene and mapping the result to the correct answer option. The oracle rejects questions with violated constraints, ambiguous answers, or insufficient visible evidence. Ground-truth answers are determined solely by the executed geometry, independently of model predictions or majority voting. The full verification procedure is described in Appendix~\ref{app:oracle}.

\paragraph{Solver.}
Following the updated Proposer and Coder, a \emph{single} data-generation pass using $\pi_t^P$ and $\pi_t^C$ produces the Solver training set $\mathcal D_t^S$. Each accepted example consists of multi-view images, a validated question from the Proposer-generated set $q_0$, and its geometry-derived ground-truth label $y^{GT}$. The Solver receives only the images and question, while scene metadata and ground-truth labels remain outside its input and are used for validation and reward computation. Validation removes geometrically inconsistent, ambiguous, visually unsupported, and duplicate examples. Unlike pseudo-label-based training, no consensus filter is needed to establish answer correctness. The resulting policy $\pi_t^S$ supplies difficulty feedback to the Proposer in the next iteration, completing the self-evolving loop. Additional details on data quality and GRPO filtering are provided in Appendix~\ref{app:data}.

\input{Table/Table1-MainResults}

\subsection{Reward Formulations}\label{sec:rewards_formulations}
We adopt role-specific rewards for task generation, scene construction, and spatial reasoning. Detailed definitions of the Proposer and Coder rewards are provided in Appendix~\ref{app:rewards}.

\paragraph{Proposer reward.}
The Proposer is rewarded for generating challenging questions, diverse spatial reasoning tasks, and realizable scene configurations:
\begin{equation}
    r^P(z)=
    \begin{cases}
        -1, & V(z)=0,\\[2pt]
        r_{\mathrm{diff}}(z)+r_{\mathrm{div}}(z)
        +r_{\mathrm{real}}(z), & V(z)=1.
    \end{cases}
    \label{eq:proposer_reward}
\end{equation}
Here, $V(z)$ indicates whether the proposal passes validation. $r_{\mathrm{diff}}$ evaluates the difficulty of validated questions based on the accuracy of the frozen Solver against geometric ground truth. $r_{\mathrm{div}}$ evaluates the number of valid questions and the coverage of spatial-relation types, encouraging both quantity and diversity. $r_{\mathrm{real}}$ measures the fraction of Coder attempts that satisfy the required scene and environment constraints. All three reward terms are normalized to $[0,1]$ and equally weighted. Detailed definitions of the Proposer reward components are provided in Appendix~\ref{app:pr}.

\paragraph{Coder reward.}
The Coder is rewarded for successfully executing the scene program and penalized for illegal object placement, unsatisfied spatial relations, and execution failure:
\begin{equation}
    r^C(c)=r_{\mathrm{exec}}(c)
          -P_{\mathrm{placement}}(c)
          -P_{\mathrm{relation}}(c)
          -P_{\mathrm{failure}}(c).
    \label{eq:coder_reward}
\end{equation}
Placement and relation violations are measured on the generated scene itself, without subsequent position correction. The two constraint penalties are equally weighted, as described in Appendix~\ref{app:cr}.

\paragraph{Solver reward.}
The Solver is trained solely with a binary correctness reward against the geometric ground truth:
\begin{equation}
    r^S(\hat{y},y^{\mathrm{GT}})
    = \mathbb{I}[\hat{y}=y^{\mathrm{GT}}],
    \label{eq:solver_reward}
\end{equation}

where $\hat{y}\in\{A,B,C,D,\bot\}$ denotes the predicted answer from the Solver, with $\bot$ indicating an unparseable response. Correct predictions receive $1$, while incorrect or unparseable predictions receive $0$. No additional format reward or consensus-based pseudo-label is used.

%% file: Table/Table1-MainResults.tex
% main result table
\begin{table*}[t!]
\centering
\caption{Main results on six spatial reasoning benchmarks. Unless specified, all benchmark categories are evaluated. \textbf{Avg.} is the mean of six benchmark-level scores. Steps denote cumulative Solver optimization steps across self-evolution iterations. Best results within each backbone group are highlighted in bold. Results are averaged over three independent runs with different random seeds.}

\label{tab:main_results}

\begingroup
\setlength{\tabcolsep}{8pt}

\resizebox{0.95\textwidth}{!}{%
\begin{tabular}{@{\hspace{6pt}}l*{7}{c}@{\hspace{6pt}}}
\toprule
\textbf{Methods} & \textbf{VSI-Bench} & \textbf{MindCube} & \textbf{MMSI}
& \textbf{ViewSpatial} & \textbf{OmniSpatial}
& \textbf{EmbSpatial} & \textbf{Avg.} \\
\midrule
\multicolumn{8}{@{}l}{\textit{Self-Evolution Methods (backbone)}} \\
\quad SpatialSkill (\textit{Qwen3.5-4B})
    & 47.25 & 47.43 & 33.00 & 43.50 & 44.03 & 77.20 & 48.74 \\    
\quad MM-Zero (\textit{Qwen3.5-4B})
    & 46.10 & 46.67 & 32.00 & 42.60 & 44.21 & 76.65 & 48.04 \\
\midrule
\multicolumn{8}{@{}l}{\textit{Qwen3.5-4B}} \\
\quad Base Model
    & 45.74 & 46.00 & 31.20 & 42.79 & 44.39 & 76.57 & 47.78 \\
\quad Spatial-Zero (Step 15, Iter 1)
    & 46.84 & 46.76 & 32.70 & 43.47 & 43.67 & 77.28 & 48.45 \\
\quad Spatial-Zero (Step 30, Iter 2)
    & 48.90 & \textbf{48.86} & 34.80 & 44.76 & 44.21 & 78.27 & 49.97 \\
\quad Spatial-Zero (Step 45, Iter 3)
    & \textbf{49.83} & 48.67 & \textbf{36.20} & \textbf{45.02} & \textbf{44.74} & \textbf{78.71} & \textbf{50.53} \\
\midrule
\multicolumn{8}{@{}l}{\textit{Qwen3.5-9B}} \\
\quad Base Model
    & 52.34 & 47.33 & 31.70 & 46.30 & 46.17 & 76.98 & 50.14 \\
\quad Spatial-Zero (Step 15, Iter 1)
    & 53.41 & 47.62 & 33.10 & 47.02 & 46.70 & 77.72 & 50.93 \\
\quad Spatial-Zero (Step 30, Iter 2)
    & 55.78 & \textbf{49.90} & 36.50 & 48.43 & 46.52 & 79.15 & 52.71 \\
\quad Spatial-Zero (Step 45, Iter 3)
    & \textbf{56.58} & 49.62 & \textbf{37.40} & \textbf{48.60} & \textbf{48.31} & \textbf{79.70} & \textbf{53.37} \\
\midrule
\multicolumn{8}{@{}l}{\textit{InternVL3.5-8B}} \\
\quad Base Model
    & 49.77 & 39.90 & 29.40 & 39.68 & 39.04 & 74.26 & 45.34 \\
\quad Spatial-Zero (Step 15, Iter 1)
    & 50.37 & 40.86 & 30.50 & 40.30 & 39.22 & 74.70 & 45.99 \\
\quad Spatial-Zero (Step 30, Iter 2)
    & 51.22 & 41.71 & 32.10 & 40.92 & \textbf{39.75} & 75.47 & 46.86 \\
\quad Spatial-Zero (Step 45, Iter 3)
    & \textbf{51.79} & \textbf{42.38} & \textbf{33.50} & \textbf{41.69} & 39.39 & \textbf{75.93} & \textbf{47.45} \\
\bottomrule
\end{tabular}%
}
\endgroup

\end{table*}

%% file: sec/4_Experiment.tex
\section{Experiment}

\subsection{Experimental Setup}\label{sec:experimental_setup}

\textbf{Implementation.}
The Proposer, Coder, and Solver are initialized from the same backbone and independently optimized using full-parameter GRPO~\cite{Deepseekmath}. Training spans three self-evolution iterations, each comprising 5, 5, and 15 optimization steps for the three roles, respectively, following R-Zero~\cite{R-Zero}. The Solver is evaluated at cumulative Steps 15, 30, and 45. Training uses ms-swift~\cite{zhao2025swift} and vLLM~\cite{kwon2023vllm}, with Blender rendering and a fixed Geometric Oracle.
\textbf{Benchmarks.}
We evaluate six benchmarks: VSI-Bench~\cite{VSI-Bench}, MindCube-Tiny~\cite{MindCube}, MMSI-Bench~\cite{MMSI-Bench}, ViewSpatial-Bench~\cite{Viewspatial-bench}, OmniSpatial-test~\cite{Omnispatial}, and EmbSpatial-Bench~\cite{embspatial-bench}, covering all task categories unless specified. For OmniSpatial, we evaluate Egocentric, Allocentric, and Hypothetical Tasks. All benchmarks follow their original evaluation protocols unless otherwise noted.
\textbf{Models and Baselines.}
We use Qwen3.5-4B/9B~\cite{qwen3.5} and InternVL3.5-8B~\cite{InternVL3.5}, comparing against pretrained models, SpatialSkill~\cite{spatialskill}, and MM-Zero~\cite{MM-Zero_EMNLP2026}. Both competing methods are reproduced with Qwen3.5-4B, retaining their original configurations except the backbone, and evaluated under the same protocol. Additional implementation and evaluation details are provided in Appendix~\ref{app:impl}.

%-------Contribution of Evolving Components
\begin{table*}[t!]
\centering
\caption{Component and Proposer reward ablations on six spatial reasoning benchmarks with Qwen3.5-4B. Avg.\ denotes the arithmetic mean of the six benchmark-level scores.}
\label{tab:ablation_components_rewards}
\begingroup
\setlength{\tabcolsep}{8pt}

\resizebox{0.95\textwidth}{!}{%
\begin{tabular}{@{\hspace{6pt}}l*{7}{c}@{\hspace{6pt}}}
\toprule
\textbf{Variant} & \textbf{VSI-Bench} & \textbf{MindCube} & \textbf{MMSI}
& \textbf{ViewSpatial} & \textbf{OmniSpatial} & \textbf{EmbSpatial}
& \textbf{Avg.} \\
\midrule
\multicolumn{8}{@{}l}{\textit{Component Ablation}} \\
\quad Solver Only & 47.64 & 47.24 & 33.40 & 43.73 & 43.92 & 77.47 & 48.90 \\
\quad Frozen Proposer & 48.20 & 47.53 & 34.34 & 44.12 & 44.12 & 77.91 & 49.37 \\
\quad Frozen Coder & 48.61 & 47.90 & 35.12 & 44.39 & 44.03 & 78.14 & 49.70 \\
\quad \textbf{Full Spatial-Zero} & \textbf{49.83} & \textbf{48.67}
& \textbf{36.20} & \textbf{45.02} & \textbf{44.74} & \textbf{78.71} & \textbf{50.53} \\
\midrule
\multicolumn{8}{@{}l}{\textit{Proposer Reward Ablation}} \\
\quad w/o $r_{\mathrm{diff}}$ & 49.02 & 47.60 & 35.10 & 44.32 & 43.91 & 77.69 & 49.61 \\
\quad w/o $r_{\mathrm{div}}$ & 48.21 & 47.89 & 35.00 & 44.05 & 43.62 & 78.03 & 49.47 \\
\quad w/o $r_{\mathrm{real}}$ & 48.92 & 48.08 & 35.49 & 44.61 & 44.10 & 78.26 & 49.91 \\
\quad \textbf{Full Spatial-Zero} & \textbf{49.83} & \textbf{48.67}
& \textbf{36.20} & \textbf{45.02} & \textbf{44.74} & \textbf{78.71} & \textbf{50.53} \\
\bottomrule
\end{tabular}%
}
\endgroup
\end{table*}

\subsection{Main Results}
As shown in Table~\ref{tab:main_results}, we evaluate Spatial-Zero on six spatial reasoning benchmarks using three VLM backbones and compare it with existing self-evolution methods under Qwen3.5-4B. The results yield four key observations:

\noindent\textbf{(1) Overall comparison.}
Spatial-Zero achieves an average score of 50.53 with Qwen3.5-4B, outperforming the SpatialSkill (48.74) and MM-Zero (48.04) baselines by 1.79 and 2.49 points, respectively. Compared with the base model (47.78), Spatial-Zero improves the average score by 2.75 points. These results demonstrate the effectiveness of Spatial-Zero relative to representative self-evolution methods under a common backbone and evaluation setting.

\noindent\textbf{(2) Performance across benchmarks.}
Spatial-Zero outperforms the strongest baseline across all six benchmarks with Qwen3.5-4B, achieving gains on MMSI (+3.20), VSI-Bench (+2.58), ViewSpatial (+1.52), EmbSpatial (+1.51), MindCube (+1.24), and OmniSpatial (+0.53). Despite the smaller gain on OmniSpatial, these consistent improvements suggest that geometry-grounded supervision benefits diverse spatial reasoning tasks.

\noindent\textbf{(3) Generalization across backbones.}
Spatial-Zero improves the average scores of Qwen3.5-4B, Qwen3.5-9B, and InternVL3.5-8B by 2.75, 3.23, and 2.11 points, respectively. After three iterations, all three models achieve improvements over their corresponding base models on all six benchmarks, with particularly notable gains on MMSI (+5.00, +5.70, and +4.10, respectively). These results demonstrate the applicability of Spatial-Zero across different scales and architectures.

\noindent\textbf{(4) Iterative self-evolution.}
The average scores increase across three iterations, reaching 50.53, 53.37, and 47.45 for Qwen3.5-4B, Qwen3.5-9B, and InternVL3.5-8B, respectively. All three models exhibit the largest gains between Iter 1 and Iter 2, followed by smaller improvements in Iter 3. While individual benchmarks exhibit occasional fluctuations, the average performance improves consistently across iterations, suggesting that self-evolution progressively enhances spatial reasoning capabilities.

\subsection{Ablation Studies}\label{sec:ablation}
We conduct ablation studies to investigate three aspects of Spatial-Zero: the contributions of the three evolving roles, the effectiveness of geometry-grounded supervision and validation, and the impact of individual Proposer reward components. Unless otherwise specified, all ablations are conducted using Qwen3.5-4B for three self-evolution iterations (45 cumulative Solver optimization steps), following the main experimental setup.

\subsubsection{Contribution of Evolving Components}
We investigate the contributions of the three evolving roles by comparing the full method with three variants: Solver Only, which freezes the Proposer and Coder while generating fresh spatial training examples in each iteration; Frozen Proposer, which updates only the Coder and Solver; and Frozen Coder, which updates only the Proposer and Solver. All variants retain the same Geometric Oracle, Solver training objective, and cumulative Solver optimization budget. As shown in~\Cref{tab:ablation_components_rewards}, the full method outperforms Solver Only, Frozen Proposer, and Frozen Coder by 1.63, 1.16, and 0.83 points in Avg., respectively. Freezing the Proposer causes a larger performance drop than freezing the Coder. Moreover, despite continuously generating fresh training examples, Solver Only achieves a lower score than the full method, highlighting the importance of evolving the task-generation policies alongside the Solver.

\subsubsection{Effect of Proposer Reward Components}
We investigate the contribution of each Proposer reward component by individually removing $r_{\mathrm{diff}}$, $r_{\mathrm{div}}$, or $r_{\mathrm{real}}$, while keeping all other training settings unchanged. As shown in~\Cref{tab:ablation_components_rewards}, removing any component reduces performance across six benchmarks. Without $r_{\mathrm{diff}}$, Avg.\ decreases by 0.92 points, indicating the benefit of difficulty-aware question generation. Removing the diversity reward $r_{\mathrm{div}}$ results in the largest average drop of 1.06 points, including a 1.62-point decrease on VSI-Bench, highlighting the importance of encouraging both the quantity and spatial-relation diversity of valid questions. Removing the realizability reward $r_{\mathrm{real}}$ reduces Avg.\ by 0.62 points, suggesting that encouraging scene configurations reliably instantiated by the Coder benefits Solver training. Overall, these results demonstrate the complementary contributions of the three reward components to Spatial-Zero.

%-------The Geometric Oracle ablations
\begin{table*}[t!]
\centering
\caption{Ablation of geometric supervision and validation using Qwen3.5-4B after three self-evolution iterations.}
\label{tab:ablation_oracle}
\begingroup
\setlength{\tabcolsep}{8pt}

\resizebox{0.95\textwidth}{!}{%
\begin{tabular}{l l l c c c c}
\toprule
\textbf{Variant} & \textbf{Label Source} & \textbf{Validation}
& \textbf{Accepted (\%)}
& \textbf{Fully Valid (\%)}
& \textbf{Agreement (\%)}
& \textbf{Avg.} \\
\midrule
\textbf{Full Oracle} & Geometry & Full
& 81.4 & 100.0 & -- & \textbf{50.53} \\
Pseudo-label & Model & Full
& 81.4 & 100.0 & 86.7 & 47.56 \\
w/o Visibility & Geometry & No visibility check
& 90.2 & 90.2 & -- & 48.78 \\
w/o Uniqueness & Geometry + Vote & No uniqueness check
& 89.9 & 90.5 & -- & 48.58 \\
\bottomrule
\end{tabular}%
}
\endgroup
\end{table*}

\subsubsection{Effect of Geometric Supervision and Validation}
We investigate the contributions of geometric supervision and sample validation through two sets of ablations. We report three diagnostic metrics computed on a fixed pool of candidate questions: Accepted Rate measures the proportion admitted by the active validation rules, Fully Valid Rate measures the proportion of accepted questions satisfying the complete Geometric Oracle criteria, and Label Agreement measures the agreement between pseudo-labels and geometry-derived ground truth on uniquely answerable questions. For geometric supervision, we replace geometry-derived ground truth with model-generated pseudo-labels while retaining full validation. As shown in~\Cref{tab:ablation_oracle}, pseudo-labels achieve 86.7\% agreement with geometric ground truth, while Avg.\ decreases by 2.97 points, highlighting the importance of reliable geometric supervision. For sample validation, we remove the visibility or uniqueness check while retaining other criteria. In the latter setting, majority voting following MM-Zero~\cite{MM-Zero_EMNLP2026} assigns a single label to non-unique questions. Both variants increase the accepted rate (90.2\% and 89.9\% vs.\ 81.4\%) but reduce the fully valid rate (90.2\% and 90.5\% vs.\ 100.0\%), with Avg.\ dropping by 1.75 and 1.95 points, respectively. These results support strict validation, while the uniqueness ablation reflects both relaxed filtering and voting-based labeling.

\noindent\textbf{Independent Visual Answerability Evaluation.}
To further assess the reliability of generated spatial experiences, we conduct an independent human evaluation on 200 questions sampled from those accepted by the full Geometric Oracle. Annotators are provided only with rendered multi-view images and multiple-choice questions, without access to scene metadata or ground-truth labels. Among these questions, 196 (98.0\%) are judged visually answerable, and 195 of the 196 answerable questions (99.5\%) agree with the geometry-derived labels. These results provide independent evidence that Oracle-validated questions are largely visually answerable and that geometric supervision is highly consistent with human judgments. Further details of the human evaluation are provided in Appendix~\ref{app:human}.

\subsection{Analysis of Self-Evolution}
We analyze Spatial-Zero through online versus static training, task evolution across iterations, and iteration scaling across model sizes. The iteration scaling analysis is presented in Appendix~\ref{app:iteration_scaling}.

\subsubsection{Online Evolution vs. Static Training}

\begin{wrapfigure}{r}{0.48\textwidth}
    \centering
    \vspace{-8pt}
    \includegraphics[width=\linewidth]{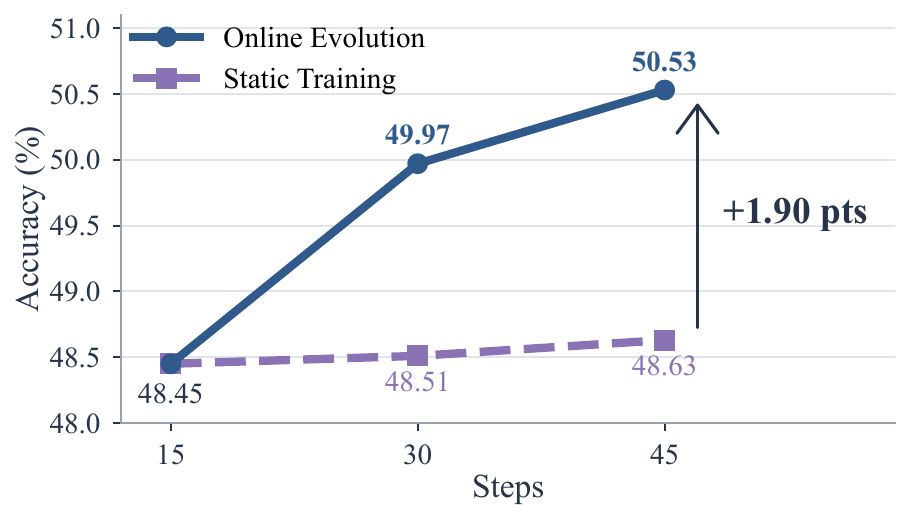}
    \caption{Comparison of online self-evolution and static-data training with Qwen3.5-4B.}
    \label{fig:online_vs_static}
    \vspace{-8pt}
\end{wrapfigure}

We investigate whether online self-evolution provides additional benefits over training on fixed generated data. Both settings share the first self-evolution iteration and the Solver checkpoint at Step 15, followed by the same Solver optimization budget and evaluation protocol. Static training reuses the validated training examples from the first iteration, while Spatial-Zero continuously updates the Proposer and Coder to generate new training tasks. As shown in~\Cref{fig:online_vs_static}, Spatial-Zero consistently improves across subsequent iterations, whereas static training exhibits only marginal gains. At Step 45, Spatial-Zero achieves 50.53 compared with 48.63 for static training, a 1.90-point improvement. Furthermore, the component ablation in~\Cref{tab:ablation_components_rewards} shows that Solver Only, which continues to generate fresh training examples using frozen Proposer and Coder policies, achieves 48.90. Together, these results highlight the benefits of continuously evolving task-generation policies alongside the Solver, rather than relying solely on fixed training data or frozen generation policies.

\subsubsection{Evolution of Generated Spatial Tasks}
\label{sec:task_evolution}

\begin{figure*}[ht]
    \centering
    \includegraphics[width= \linewidth]{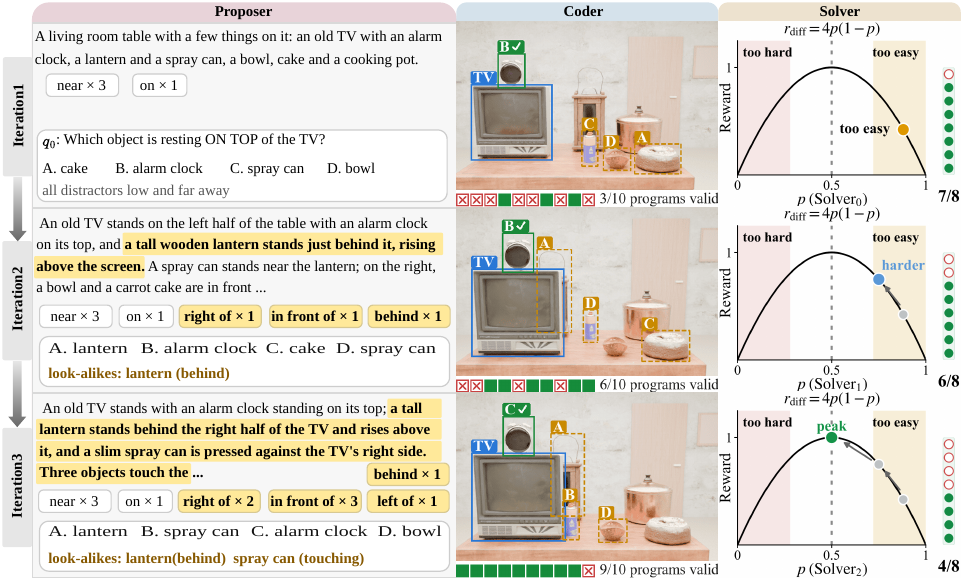}
    \caption{Generated tasks become harder as scene realization improves. Columns show a representative task across three iterations. Green boxes mark the correct object; orange dashed boxes mark distractors.}
    \label{fig:task_evolution}
\end{figure*}

Figure~\ref{fig:task_evolution} follows a representative task through three self-evolution iterations. The question remains ``Which object is resting on top of the TV?'', and the Geometric Oracle consistently identifies the alarm clock as the answer. The Proposer first places the distractors away from the television, then introduces a lantern behind and above it and a spray can touching its side. It also specifies a broader range of spatial relations. As these proposals evolve, valid Coder programs increase from 3/10 to 6/10 and 9/10. The frozen Solver's accuracy on the question falls from 7/8 to 6/8 and 4/8, bringing the task to the maximum of the difficulty reward $r_{\mathrm{diff}}=4p(1-p)$ at $p=0.5$. This illustrates how the Proposer can create a more challenging question while the Coder improves its ability to realize the proposed scene.

%% file: sec/2_RelatedWork.tex
\section{Related Work}

\textbf{Self-Evolution in Vision-Language Models.}
Self-evolution updates a model using tasks and feedback generated within its own training loop~\cite{guo2026seeing}. In language-only reasoning, R-Zero co-evolves a Challenger and a Solver, while Absolute Zero uses code execution to validate self-proposed tasks and answers~\cite{R-Zero,Absolute-Zero}.
Extending this paradigm to vision-language models (VLMs) introduces a requirement: the loop must obtain visual inputs. V-Zero and VisPlay co-evolve question generation and solving over unlabeled images, removing answer annotations while retaining external image collections~\cite{V-Zero,VisPlay}.
Trinity employs an exponential moving average policy to assess visual grounding and answer correctness for generated questions~\cite{lee2026trinity}.
MM-Zero removes the external image dependency by introducing a Coder that renders structured visual concepts through executable programs~\cite{MM-Zero_EMNLP2026}. However, its Solver relies on majority-voted pseudo-labels for challenging questions rather than answers verified against an independent scene state.
Spatial-Zero builds on programmatic generation but executes 3D scene programs whose geometry determines question validity and ground-truth answers, connecting task evolution with correctness-grounded feedback.

\textbf{Spatial Reasoning in Vision-Language Models.}
Spatial benchmarks assess video-based recall, 3D object relations, and multi-view reasoning, including VSI-Bench and 3DSRBench~\cite{VSI-Bench,3DSRBench}, as well as MMSI-Bench and MindCube~\cite{MMSI-Bench,MindCube}.
SpatialVLM, SpatialRGPT, and SPAR-7M derive spatial supervision from existing images or 3D scenes~\cite{SpatialVLM,SpatialRGPT,zhang2026flatland}, while Spatial-SSRL constructs verifiable tasks from RGB or RGB-D observations~\cite{Spatial-SSRL}.
Programmatic approaches include CLEVR, which generates scenes and executes question programs~\cite{CLEVR_2017_CVPR}, Spatial-DISE, which synthesizes verifiable 3D tasks in Blender, and SPRITE, which derives instruction-tuning data from simulator metadata~\cite{Spatial-dise, SPRITE_CVPR26Findings}. Feedback-based methods include Spatial Memory Agent and SpatialSkill, which accumulate experience or skills for frozen VLMs~\cite{SMA,spatialskill}, SAGE, which adapts geometric and linguistic transformations, and Spatial-OPSD, which distills privileged spatial priors~\cite{SAGE,Spatial-OPSD}.
SpatialEvo verifies questions against existing 3D scenes while jointly optimizing question generation and solving~\cite{Spatialevo}.
In contrast, Spatial-Zero generates 3D configurations and questions within its training loop, using geometry for reliable supervision and verification and Solver feedback to guide subsequent task generation.

%% file: sec/5_Conclusion.tex
\section{Conclusion}

In this paper, we introduced Spatial-Zero, a zero-data self-evolving framework that couples autonomous 3D scene generation with deterministic geometric supervision through three iteratively optimized roles: the Proposer generates spatial configurations and questions, the Coder constructs executable 3D scenes, and the Solver learns from geometry-verified supervision. A deterministic Geometric Oracle validates generated tasks and derives ground-truth answers, enabling spatial experiences and reasoning capabilities to co-evolve without external spatial training data or model-voted pseudo-labels. Experiments across six spatial reasoning benchmarks and three VLM backbones demonstrate consistent improvements over base models and existing self-evolution methods, while further analyses confirm the benefits of geometric supervision and iterative co-evolution across model scales and architectures.

%% file: sec/X_Appendix.tex
\newpage

\section*{Appendix Overview}
\addcontentsline{toc}{section}{Appendix Overview}
\noindent\textbf{Appendix A: Supplementary Methodology.}
Complements main-paper \S\S2.2--2.4 with proposal validation, Coder scene construction, Geometric Oracle verification, Solver data filtering, detailed reward functions, and sequential optimization.

\noindent\textbf{Appendix B: Prompt Interfaces.}
Complements \S2.3 with the input instructions and output formats for the Proposer, Coder, and Solver.

\noindent\textbf{Appendix C: Implementation and Evaluation Details.}
Complements \S3.1 with training configurations, scene assets and rendering, computational resources and training efficiency, benchmark evaluation and score aggregation, and baseline reproduction.

\noindent\textbf{Appendix D: Additional Ablation Details.}
Complements \S3.3 with an independent human evaluation of visual answerability, agreement with the Geometric Oracle, and inter-annotator consistency.

\noindent\textbf{Appendix E: Additional Self-Evolution Analysis.}
Complements \S3.4 with iteration scaling across model sizes across three VLM backbones.

\appendix

%%%%%------------------------------Appendix A

\section{Supplementary Methodology}\label{app:method}
\mainsec{2.2--2.4}
This section details proposal validation, program realization, geometric verification, data filtering, and role-specific optimization. We distinguish proposal-level failures from invalid questions and separate static program checks from verification of rendered scenes. The Proposer and Coder reward definitions expand main-paper Eqs.~(3) and (4); the Solver correctness reward is defined in main-paper Eq.~(5).

\subsection{Proposal Representation and Validation}\label{app:proposal}
\paragraph{Proposal representation.}
Following main-paper \S2.3, a proposal is represented as $z=(e,\mathcal O,\mathcal C,q_0)$, where $e$ specifies an indoor or outdoor environment, $\mathcal O$ is the target-object set, $\mathcal C$ contains mandatory scene and spatial constraints, and $q_0=\{q_{0,j}\}_{j=1}^{J}$ contains four-option questions without answer labels. The Proposer selects these specifications, whereas the Coder determines the realized object geometry. Ground-truth labels are derived only after scene realization.

\paragraph{Proposal-level and question-level validity.}
Preliminary checks verify the proposal format, supported environments and objects, object references, option structure, and feasibility of required layouts and relations. Subsequent scene verification checks object identity, placement, collision, support, and mandatory spatial constraints. A proposal is invalid if the mandatory preliminary checks fail or no attempted scene realization passes the required scene and question checks within the evaluation budget; such a proposal receives $r^P(z)=-1$. For an accepted scene, geometrically ambiguous, duplicated, or visually unsupported individual questions are discarded. A failed individual question does not require rewriting the other valid questions. The validation status of a question is therefore distinct from the validity of the proposal as a whole.

\subsection{Coder Scene Construction and Execution}\label{app:coder}
\paragraph{Program sampling and static verification.}
For each proposal, the Coder samples $N$ candidate Blender programs $\{c_n\}_{n=1}^{N}$. Candidate programs undergo deterministic interface checks and static evaluation of required objects, legal placement, and spatial constraints. The checks produce an executability indicator $u_n$, a legal-placement rate $L_n$, and a relation-satisfaction rate $F_n$, as defined in~\Cref{app:cr}. Missing required objects are counted as placement violations, and their associated relations are treated as unsatisfied. Coder optimization uses the static program score; the generated object positions are not repaired to improve this score.

\paragraph{Render selection and scene verification.}
Statically eligible candidates are considered in sampling order under a bounded rendering budget. For each attempted candidate, the normalized Blender program is executed to produce multiview observations $I$ and scene metadata $M$. The executed scene is checked again for required objects, collisions, supports, and relations. Only a candidate passing these checks and the question-level Oracle verification is accepted. A subsequent ordinary execution failure invalidates that candidate; failure of a scene or question check removes its eligibility without changing its geometry. This separates inexpensive checks over candidate programs from the more expensive rendered-scene validation. The scene-generation configuration samples $N=10$ candidates and attempts to render at most two statically eligible candidates. Rendering settings and asset sources are described in~\Cref{app:render}.

\subsection{Geometric Oracle and Question Verification}\label{app:oracle}
\paragraph{Verification workflow.}
The non-trainable Geometric Oracle evaluates a proposed question using executed scene geometry and the corresponding rendered observations. It first verifies question syntax and object references, then applies the appropriate geometric predicate and answer-uniqueness check, and finally verifies visual evidence. For a realized program $c$, the retained question set is
\begin{equation}
\begin{aligned}
\Qvalid(z,c)=\operatorname{Dedup}\!\bigl(\{q\in q_0:\ &\\[-1pt]
\operatorname{Valid}(q,M(c),I(c),z)=1\}\bigr).
\end{aligned}
\label{eq:app_filter}
\end{equation}
The transition from $q_0$ to $\Qvalid$ is filtering, not question rewriting or an additional question-generation stage. For each retained question, the ground-truth answer is obtained from the realized scene metadata:
\begin{equation}
\GT=\Omega(M(c),q).
\label{eq:app_gt}
\end{equation}
For clarity, $\Qvalid(z)$ in subsequent equations abbreviates the validated set for the selected realization $c$. No Solver predictions or majority-voted answers are used to determine $\GT$.

\paragraph{Verification algorithm.}
Algorithm~\ref{alg:sz_oracle} summarizes how the Oracle filters the
Proposer-generated questions and derives geometry-based labels. Scene
constraints are checked before question validation; passing the geometry
check alone does not guarantee sufficient visual evidence.

\begin{algorithm}[t]
\caption{Geometric Oracle verification.}
\label{alg:sz_oracle}
\footnotesize
\begin{algorithmic}[1]
\Require Proposal $z$, program $c$, images $I$, metadata $M$, questions $q_0$
\Ensure Validated question--label pairs $\mathcal A$, or failure
\If{$\neg\Call{SceneValid}{c,M,z}$}
    \State \Return failure
\EndIf
\State $\mathcal A\gets\emptyset$
\ForAll{$q\in q_0$}
    \If{$\Call{InvalidReferencesOrOptions}{q,M}$}
        \State \textbf{continue}
    \EndIf
    \If{$\Call{DistanceQuestion}{q}$}
        \State $y\gets\Call{UniqueDistanceWinner}{M,q}$
    \Else
        \State $s\gets\Call{PredicateStates}{M,q}$
        \State $y\gets\Call{UniqueTrueOption}{s}$
    \EndIf
    \If{$y=\bot$ \textbf{or} $\neg\Call{Visible}{q,I,M}$}
        \State \textbf{continue}
    \EndIf
    \If{$\Call{Duplicate}{q,\mathcal A}$}
        \State \textbf{continue}
    \EndIf
    \State $\mathcal A\gets\mathcal A\cup\{(q,y)\}$
\EndFor
\State \Return $\mathcal A$
\end{algorithmic}
\end{algorithm}

\paragraph{Geometric predicates and ambiguity margins.}
Geometric predicates operate on world-coordinate axis-aligned bounding-box (AABB) geometry, with camera-frame transformations applied to viewpoint-dependent relations. A scene scale factor $k$ is set to $0.35$ for tabletop configurations and to $1$ otherwise. Only thresholds explicitly multiplied by $k$ vary with scene scale. Non-distance predicates can return \emph{true}, \emph{false}, or \emph{ambiguous}; a question is accepted only if exactly one option is true and all other options are false. \Cref{tab:oracle_rules} summarizes the geometric rules.
\begin{table*}[t]
\centering\small
\caption{Geometry-based question predicates. AABB centers are used for distance and directional relations. The scale factor is $k=0.35$ for tabletop scenes and $k=1$ otherwise. Cases outside the stated true/false margins are treated as ambiguous.}
\label{tab:oracle_rules}
\renewcommand{\arraystretch}{1.13}
\begin{tabularx}{0.96\textwidth}{p{3.15cm}X}
\toprule
Question type & Verification condition \\
\midrule
Nearest / farthest & Let $d_1$ and $d_2$ be the winner and runner-up distances between AABB centers. Require $\Delta=|d_1-d_2|\geq0.30k$\,m and $\Delta/\max(d_1,10^{-6})\geq0.10$. \\
Left / right; front / behind & Transform AABB centers to the specified camera frame. A signed direction-oriented lateral or depth difference $\geq0.30k$\,m is true; a difference $\leq-0.30k$\,m is false. Camera depth uses $d=-z_{\mathrm{cam}}$. \\
Higher / lower & Use the AABB bottom-face height $z_{\min}$. A direction-oriented difference $\geq0.10k$\,m is true; a difference $\leq0.02$\,m is false. \\
Between two anchors & For anchor-center segment length $\ell\geq0.50k$\,m, define $\tau=\max(0.35k,0.08\ell)$. True if normalized projection $t\in[0.15,0.85]$ and perpendicular distance $d_\perp\leq\tau$. False if $t<-0.05$, $t>1.05$, or $d_\perp>\tau+0.30k$\,m. \\
On / supported by & For a candidate and a support-part AABB, define $\rho$ as the XY overlap divided by the candidate XY area and $g=z_{\min,\mathrm{candidate}}-z_{\max,\mathrm{part}}$. True if any support part has $\rho\geq0.35$ and $|g|\leq0.08$\,m. False if every part has $\rho\leq0.05$ or $|g|\geq0.30$\,m. \\
\bottomrule
\end{tabularx}
\end{table*}

\paragraph{Scene-level checks.}
Collision checks use AABB overlap with a $0.02$\,m tolerance during static evaluation and $0.04$\,m after scene execution. A conventional support check requires the vertical gap between the object bottom and the support surface to be at most $0.06$\,m in absolute value, with the object XY center falling within the support surface. For explicit placement constraints, world-axis direction checks use a fixed $0.30$\,m margin. These scene-level checks are distinct from question predicates, which may use camera-relative coordinates and scale-dependent margins.

\paragraph{Visibility and answer uniqueness.}
Visibility is measured separately for each object and each rendered view, using ray tests over sampled mesh vertices and face centers (up to 160 samples per object). The visible ratio divides the number of in-frame, unoccluded samples by the total number of sampled points, including out-of-frame samples. The projected pixel extent is the larger side of the bounding box of visible projected samples. A referenced object must satisfy both a visible ratio of at least $0.10$ and a pixel extent of at least $10$ pixels. Camera-relative questions require every referenced object to pass these checks in the specified view; other question types allow each referenced object to pass in any available view. A geometrically unique answer can therefore still be rejected when image evidence is insufficient. Figure~2 of the main paper illustrates duplicate-option rejection, missing-reference rejection, and an accepted distance question.

\subsection{Solver Data Generation and Quality Filtering}\label{app:data}
\paragraph{Verified data construction.}
After updating the Proposer and Coder, a single data-generation pass builds the Solver training set
\begin{equation}
\mathcal D_t^S=\left\{(I,q,\GT):
q\in\Qvalid(z),\ \GT=\Omega(M,q)\right\}.
\label{eq:app_dataset}
\end{equation}
Accepted training examples must have complete multiview images, valid question references, sufficient visual evidence, and geometry-derived answer labels. Duplicate examples are removed. The Solver receives only the images and question, not object geometry or ground-truth labels. No model-accuracy threshold is applied to select the generated training questions.

\paragraph{GRPO learning-signal filtering.}
Sample validity and optimization-time filtering are separate. In the Solver GRPO configuration, each question produces $G=8$ sampled responses. Groups with identical binary correctness rewards have zero relative reward variance. Dynamic sampling attempts to replace these groups by drawing additional prompts, with a maximum of three resampling rounds. If the retry budget does not supply enough nonzero-variance groups, the training implementation falls back to the original rollout batch. The affected questions remain in the training data; they are not relabeled or declared geometrically invalid. Dynamic sampling is disabled for the Proposer and Coder, whose GRPO normalization settings follow main-paper \S2.1.

\subsection{Detailed Reward Functions}\label{app:rewards}
\subsubsection{Proposer Reward}\label{app:pr}
\paragraph{Difficulty.}
For a validated question $q$, the frozen previous Solver $\pi_{t-1}^S$ generates $K$ responses, yielding accuracy against geometric ground truth
\begin{equation}
p(q)=\frac1K\sum_{k=1}^{K}\ind\!\left[
\hat y_q^{(k)}=\Omega(M,q)\right].
\end{equation}
The difficulty-shaping function $D(p)=4p(1-p)$ favors intermediate difficulty and reaches its maximum at $p=0.5$. Under the question-set formulation in the main paper, difficulty is aggregated over validated questions:
\begin{equation}
r_{\mathrm{diff}}(z)=\frac{1}{|\Qvalid(z)|}
\sum_{q\in\Qvalid(z)}4p(q)[1-p(q)].
\label{eq:app_diff}
\end{equation}
The expression is evaluated only when the validated set is nonempty. A question with zero Solver accuracy may still be geometrically valid.

\paragraph{Valid-question quantity and relation coverage.}
Let $B_q$ denote the maximum number of questions proposed and $\mathcal T$ the supported spatial-relation families. The normalized quantities are
\begin{align}
r_{\mathrm{count}}(z)&=\frac{|\Qvalid(z)|}{B_q},\\
r_{\mathrm{cover}}(z)&=\frac{|\mathcal T(\Qvalid(z))|}{|\mathcal T|},\\
r_{\mathrm{div}}(z)&=\frac12\left(r_{\mathrm{count}}(z)
+r_{\mathrm{cover}}(z)\right).
\label{eq:app_div}
\end{align}
The factor $1/2$ assigns equal weight to the number of valid questions and spatial-relation coverage. Because both terms are normalized to $[0,1]$, their arithmetic mean also lies in $[0,1]$, keeping $r_{\mathrm{div}}$ on the same scale as the other Proposer reward components. The count and coverage are computed over validated questions in $q_0$, without an additional question-generation stage.

\paragraph{Scene realizability.}
Let $v_n\in\{0,1\}$ indicate whether candidate $c_n$ passes the required realization checks. Candidates are first evaluated statically; candidates selected for rendering have their eligibility updated according to execution and rendered-scene verification. Candidates not selected for rendering retain their static eligibility. The realizability term is therefore a hybrid static-and-render-verified fraction rather than a render-based success rate for every candidate:
\begin{equation}
r_{\mathrm{real}}(z)=\frac1N\sum_{n=1}^{N}v_n.
\label{eq:app_real}
\end{equation}
The complete Proposer reward follows main-paper Eq.~(3):
\begin{equation}
r^P(z)=\begin{cases}
-1,&V(z)=0,\\
r_{\mathrm{diff}}(z)+r_{\mathrm{div}}(z)+r_{\mathrm{real}}(z),&V(z)=1.
\end{cases}
\label{eq:app_proposer_total}
\end{equation}
All three components are normalized to $[0,1]$ and have equal weight in the valid-proposal case.

\subsubsection{Coder Reward}\label{app:cr}
The Coder reward is computed from static program evaluation, without correcting object positions after scoring. Let $u(c)\in\{0,1\}$ indicate that a program passes deterministic interface and static executability checks. Let $L(c)\in[0,1]$ denote the fraction of required objects with legal placement, and $F(c)\in[0,1]$ the fraction of satisfied required spatial constraints. The constraint set includes explicit placements and the conditions associated with the proposed question. The additive terms of main-paper Eq.~(4) are
\begin{align}
R_{\mathrm{exec}}(c)&=u(c),\\
P_{\mathrm{placement}}(c)&=\tfrac12u(c)[1-L(c)],\\
P_{\mathrm{relation}}(c)&=\tfrac12u(c)[1-F(c)],\\
P_{\mathrm{failure}}(c)&=1-u(c).
\end{align}
The two penalty terms have the same factor $1/2$ to balance legal object placement and spatial-relation satisfaction. In particular, for a statically executable program, the penalties sum to the mean unsatisfied fraction, so the reward reduces to the mean satisfaction rate:
\begin{equation}
r^C(c)=\begin{cases}
\tfrac12\left[L(c)+F(c)\right],&u(c)=1,\\
-1,&u(c)=0.
\end{cases}
\label{eq:app_coder_reward}
\end{equation}
Consequently, executable programs receive rewards in $[0,1]$, whereas failures receive $-1$. Missing required objects lower $L(c)$ and make the associated relation constraints unsatisfied. This equal weighting is a design choice, rather than a requirement of GRPO.

\subsection{Sequential Optimization}\label{app:seq}
The three roles are updated sequentially as $\pi_t^P\rightarrow\pi_t^C\rightarrow\pi_t^S$, with each role initialized from its own preceding checkpoint. During the Proposer phase, the frozen previous Coder $\pi_{t-1}^C$ supplies realizability checks and the frozen previous Solver $\pi_{t-1}^S$ supplies geometry-grounded difficulty feedback. During the Coder phase, proposals are generated by the updated and frozen Proposer $\pi_t^P$, while the Coder is optimized using the static reward in~\Cref{app:cr}. Finally, the updated Proposer and Coder generate a fresh, Oracle-verified training set for the Solver in a single pass. The Solver is then optimized from $\pi_{t-1}^S$. No Solver rollouts are required during this final data-generation pass. The Oracle is deterministic and remains fixed throughout all phases. The Proposer and Solver use group-level GRPO reward normalization, whereas the Coder uses batch-level normalization, consistent with main-paper \S2.1. The optimization budgets and implementation settings are given in Appendix C.

\paragraph{Sequential training algorithm.}
Algorithm~\ref{alg:sz_training} summarizes the policy updates and the single verified-data generation pass in each iteration. The helper \texttt{RealizeAndVerify} performs candidate screening, bounded rendering, and Algorithm~\ref{alg:sz_oracle} question validation, as described in~\Cref{app:coder,app:oracle}. The Proposer phase uses frozen previous-iteration Coder and Solver policies for feedback; the Coder and Solver phases use the updated upstream policies only as frozen data generators. The pseudocode follows the question-set formulation used in the main paper. The \texttt{TrainProposer}, \texttt{TrainCoder}, and \texttt{TrainSolver} routines denote the respective GRPO optimization stages rather than additional algorithms.

\begin{algorithm}[t]
\caption{Sequential Spatial-Zero self-evolution.}
\label{alg:sz_training}
\footnotesize
\begin{algorithmic}[1]
\Require Base VLM $\pi_0$, number of iterations $T$
\Ensure Updated policies $(\pi_T^P,\pi_T^C,\pi_T^S)$
\State Initialize $\pi_0^P,\pi_0^C,\pi_0^S$ from $\pi_0$
\State Keep Geometric Oracle $\Omega$ fixed
\For{$t=1,\ldots,T$}
    \State \textbf{Proposer:} freeze $\pi_{t-1}^C$ and $\pi_{t-1}^S$
    \State $\pi_t^P\gets\Call{TrainProposer}{\pi_{t-1}^P;\pi_{t-1}^C,\pi_{t-1}^S,\Omega}$
    \State \textbf{Coder:} freeze updated $\pi_t^P$
    \State $\mathcal D_t^C\gets\Call{ValidProposals}{\pi_t^P}$
    \State $\pi_t^C\gets\Call{TrainCoder}{\pi_{t-1}^C,\mathcal D_t^C,r^C}$
    \State \textbf{Solver data:} freeze $\pi_t^P$ and $\pi_t^C$
    \State $\mathcal D_t^S\gets\emptyset$
    \ForAll{$z$ sampled from $\pi_t^P$}
        \State $(I,\mathcal A)\gets\Call{RealizeAndVerify}{z,\pi_t^C,\Omega}$
        \If{realization succeeds}
            \State $\mathcal D_t^S\gets\mathcal D_t^S\cup\{(I,q,y):(q,y)\in\mathcal A\}$
        \EndIf
    \EndFor
    \State \textbf{Solver:} train with Oracle labels
    \State $\pi_t^S\gets\Call{TrainSolver}{\pi_{t-1}^S,\mathcal D_t^S,\Omega}$
\EndFor
\State \Return $(\pi_T^P,\pi_T^C,\pi_T^S)$
\end{algorithmic}
\end{algorithm}

%%%%%------------------------------Appendix B
% Appendix B: aligned with main-paper Section 2.3.

\section{Prompt Interfaces}\label{app:prompts}
\mainsec{2.3}
We provide the input instructions and output formats for the Proposer, Coder, and Solver. The Proposer specifies spatial configurations and questions, the Coder constructs executable scenes, and the Solver answers validated questions from rendered multiview images.

\subsection{Proposer Interface}\label{app:prompt_proposer}
The Proposer generates a proposal $z=(e,\mathcal O,\mathcal C,q_0)$ consisting of an indoor or outdoor environment, target objects, mandatory spatial constraints, and a set of multiple-choice questions $q_0$. Each question contains four answer options without a ground-truth label. Object references must be consistent with the proposed scene, and camera-relative questions must specify the relevant viewpoint. The ground-truth labels are determined by the Geometric Oracle from the executed scene geometry (\Cref{app:oracle}).

\begin{promptbox}{Proposer Prompt}
INPUT:
  Available indoor/outdoor environments
  Supported objects, assets, and spatial relations
  Mandatory scene and question constraints
  Available viewpoints for camera-relative questions

TASK:
  Select one environment and the target objects.
  Specify mandatory object placements and relations.
  Generate a set of spatial multiple-choice questions.
  Provide four distinct options (A, B, C, D) per question.
  Use valid object references and, when required,
  identify the relevant camera viewpoint.
  Do not provide ground-truth answer labels.

OUTPUT:
  {
    "environment": {"preset": ...},
    "description": ...,
    "objects": [...],
    "constraints": [...],
    "questions": [
      {"question": ..., "type": ...,
       "options": {"A": ..., "B": ...,
                   "C": ..., "D": ...},
       "reference": ..., "view": ...},
      ...
    ]
  }
  Include reference and view only when required.
\end{promptbox}

\subsection{Coder Interface}\label{app:prompt_coder}
The Coder receives the environment specification, target objects, required spatial relations, and proposed questions. It produces an executable Blender Python program that assigns object positions and attributes to realize the proposal. The program is evaluated for executability, legal placement, and spatial-constraint satisfaction without correcting the scene layout after geometric verification. The Geometric Oracle independently validates the realized scene and questions, including visibility and answer uniqueness (\Cref{app:coder,app:oracle}).

\begin{promptbox}{Coder Prompt}
INPUT:
  Selected environment and scene specification
  Target object names, shapes/assets, and dimensions
  Mandatory placements and spatial constraints
  Proposed questions and required scene relations

TASK:
  Construct the scene using Blender Python.
  Place the objects to satisfy the required relations.
  Create each target with the provided constructors.
  Output an executable scene-construction program.

OUTPUT:
  Python source defining a build_scene() function:
      def build_scene():
          import bpy
          ...
  Available object constructors:
      add_primitive(name, shape, color,
          position=(x, y, z_base),
          dimensions=(dx, dy, dz))
      place_asset(name, asset_class,
          position=(x, y, z_base), yaw=angle)
  Specify object positions with numeric arguments.
  Do not generate cameras, lighting, or answer labels.

VALIDATION:
  Program: execution and placement/relation checks
  Oracle: scene and question verification,
          visibility, uniqueness, geometric labels
\end{promptbox}

\subsection{Solver Interface}\label{app:prompt_solver}
The Solver receives four views of the same rendered scene and one validated multiple-choice question. Questions involving viewpoint-dependent relations identify the relevant image. The Solver has no access to the scene metadata, camera parameters, or ground-truth labels. Its output is parsed as one of the four answer options; as specified in the main paper, the reward depends only on correctness, with no auxiliary format reward.

\begin{promptbox}{Solver Prompt}
INPUT:
  [Image 1] [Image 2] [Image 3] [Image 4]
  (Four views of the same static 3D scene)

  QUESTION: <validated spatial question>
  A. <option A>
  B. <option B>
  C. <option C>
  D. <option D>

TASK:
  Reason about the spatial configuration in 3D.
  Use the designated image for camera-relative questions.
  Select one answer from A, B, C, or D.

OUTPUT:
  Answer: X
  (X is A, B, C, or D)
\end{promptbox}

%%%%%------------------------------Appendix C

\section{Implementation and Evaluation Details}\label{app:impl}
\mainsec{3.1}
This section provides the optimization settings, scene construction and rendering pipeline, computational setup, and evaluation protocol used in Spatial-Zero.

\subsection{Training Configuration}\label{app:training}
\paragraph{Role optimization.}
The Proposer, Coder, and Solver are initialized from the same VLM backbone but maintain separate model parameters. Each role is optimized independently using full-parameter Group Relative Policy Optimization (GRPO), with the other roles held fixed during its update. At iteration $t$, the training order is Proposer $\rightarrow$ Coder $\rightarrow$ Solver, and each role is initialized from its own checkpoint at iteration $t-1$. The Geometric Oracle remains fixed throughout training. We use ms-swift for optimization, vLLM for policy rollouts, and Blender for scene rendering.

Training proceeds for three self-evolution iterations, with \textbf{5 Proposer, 5 Coder, and 15 Solver optimization steps per iteration}. Accordingly, Steps 15, 30, and 45 in the main paper denote \emph{cumulative Solver steps}, rather than the sum of updates across all three roles. The experiments use eight NVIDIA A800 GPUs. The optimization and rollout settings are summarized in~\Cref{tab:hyper}.

\begin{table*}[t]
\centering\small
\caption{Optimization, rollout, and scene-generation settings. Optimization steps are specified per role and per self-evolution iteration; the reported step counts in the main paper refer to cumulative Solver steps.}
\label{tab:hyper}
\setlength{\tabcolsep}{6pt}
\begin{tabular}{lccc}
\toprule
Parameter & Proposer & Coder & Solver \\
\midrule
Optimizer & AdamW & AdamW & AdamW \\
Learning rate (constant) & $1\times10^{-6}$ & $1\times10^{-6}$ & $1\times10^{-6}$ \\
KL coefficient $\beta$ & 0.01 & 0.01 & 0.01 \\
Clip range $(\epsilon_{\mathrm{low}},\epsilon_{\mathrm{high}})$ & $(0.2,\,0.28)$ & $(0.2,\,0.28)$ & $(0.2,\,0.28)$ \\
Rollouts per prompt $G$ & 8 & 8 & 8 \\
Prompt groups per optimization step & 8 & 16 & 32 \\
\textbf{Optimization steps per iteration} & \textbf{5} & \textbf{5} & \textbf{15} \\
Temperature / top-$p$ & 0.8 / 0.95 & 1.0 / 1.0 & 1.0 / 1.0 \\
Maximum completion tokens & 2,048 & 4,096 & 512 \\
Advantage normalization & Group & Batch & Group \\
Dynamic rollout sampling & Disabled & Disabled & Enabled \\
\midrule
\multicolumn{4}{l}{\emph{Scene generation and implementation}}\\
Coder programs per proposal $N$ & \multicolumn{3}{c}{10} \\
Frozen-Solver responses per valid question $K$ & \multicolumn{3}{c}{8} \\
Views / image resolution & \multicolumn{3}{c}{4 / $512\times512$} \\
Renderer / samples per pixel & \multicolumn{3}{c}{Blender Cycles / 64} \\
Optimization / rollout system & \multicolumn{3}{c}{ms-swift / vLLM} \\
Distributed training / precision & \multicolumn{3}{c}{DeepSpeed ZeRO-3 / bfloat16} \\
Hardware & \multicolumn{3}{c}{$8\times$ NVIDIA A800 GPUs} \\
\bottomrule
\end{tabular}
\end{table*}

\paragraph{GRPO advantage normalization.}
As specified in the main paper, the Proposer and Solver normalize relative advantages using group-level reward standard deviations, whereas the Coder uses a batch-level standard deviation. For Solver optimization, dynamic sampling prioritizes rollout groups with nonuniform binary correctness rewards. This optimization-stage mechanism is distinct from the geometric validity checks used to construct the training set (\Cref{app:data}).

\begin{table*}[t]
\centering\small
\caption{Training cost (hours) of Spatial-Zero with the 5/5/15 Proposer/Coder/Solver optimization schedule on eight NVIDIA A800 GPUs. The single-iteration breakdown and the cumulative three-iteration cost are reported separately.}
\label{tab:app_runtime}
\setlength{\tabcolsep}{12pt}
\begin{tabular}{lrrrr}
\toprule
 & \multicolumn{3}{c}{Single iteration (h)} & Three iterations (h) \\
\cmidrule(lr){2-4}\cmidrule(l){5-5}
Backbone & Data Gen & GRPO & Total & Total \\
\midrule
Qwen3.5-4B      & 1.50 & 2.28 & 3.78 & 12.02 \\
Qwen3.5-9B      & 1.43 & 3.98 & 5.42 & 16.69 \\
InternVL3.5-8B  & 2.15 & 9.55 & 11.71 & 35.81 \\
\bottomrule
\end{tabular}
\end{table*}

\begin{wraptable}{r}{0.48\textwidth}
    \centering
    \vspace{-8pt}
    \small
    \caption{Peak GRPO trainer-reported GPU memory (GiB).}
    \label{tab:app_memory}
    \setlength{\tabcolsep}{3pt}
    \begin{tabular}{@{}lccc@{}}
        \toprule
        Backbone & Proposer & Coder & Solver \\
        \midrule
        Qwen3.5-4B & 53.62 & 46.85 & 40.28 \\
        Qwen3.5-9B & 75.20 & 58.96 & 54.00 \\
        \bottomrule
    \end{tabular}
    \vspace{-8pt}
\end{wraptable}

\paragraph{GPU memory.}
\Cref{tab:app_memory} reports the maximum GPU-memory values recorded by the GRPO training processes for each role in the Qwen3.5 experiments.

\begin{table*}[t]
\centering\small
\caption{Evaluation coverage and benchmark-level aggregation. The final Avg. is the arithmetic mean of six benchmark-level scores.}
\label{tab:eval_protocol}
\begin{tabularx}{0.95\textwidth}{lXl}
\toprule
Benchmark & Evaluated tasks or categories & Aggregation \\
\midrule
VSI-Bench & All eight categories & Eight-category mean \\
MindCube-Tiny & All three categories & Full benchmark \\
MMSI-Bench & Overall & Overall accuracy \\
ViewSpatial-Bench & Cam-Rel, Per-Rel, Per-Sce-Sim & Three-category macro average \\
OmniSpatial-test & Egocentric (102), Allocentric (376), Hypothetical (83) & Micro average over 561 questions \\
EmbSpatial-Bench & All six categories & Full benchmark \\
\bottomrule
\end{tabularx}
\end{table*}

\subsection{Scene Assets and Environment Templates}\label{app:scene_assets}
\paragraph{Procedural environments.}
Spatial-Zero constructs indoor and outdoor scenes from reusable environment specifications rather than retrieving externally annotated spatial question--answer examples. Indoor configurations include studios, living rooms, bedrooms, offices, kitchens, classrooms, and warehouses; outdoor configurations include parks, streets, gardens, campsites, lakesides, cafe terraces, and loading yards. Environment specifications define scene boundaries, background geometry, semantic objects, and support surfaces. Layouts of walls, floors, furniture, and outdoor elements are assembled procedurally, while the Proposer chooses the environment for each proposal instead of following a fixed environment assignment across iterations.

\paragraph{Target objects and visual assets.}
Each proposed scene contains 4--8 target objects. Targets are constructed from geometric primitives, such as cubes, spheres, cylinders, and cones, or selected from an object-asset catalog. The catalog provides object classes, dimensions, and geometry information used to enforce placement and support constraints. The asset pipeline supports individual 3D object models, physically based rendering (PBR) materials, and high-dynamic-range environment lighting; the available resources include CC0 object models from Poly Haven and PBR textures from Poly Haven and ambientCG. These visual resources are distinguished from spatial training annotations: the questions used to optimize the Solver and their ground-truth labels are derived from generated scenes and verified geometry, not from an externally supplied spatial question--answer dataset.

\subsection{Blender-Based Scene Generation and Rendering}\label{app:render}
\paragraph{Scene realization.}
Given a Proposer output, the Coder generates a Blender Python program specifying the positions and attributes of the target objects. The scene-construction interface provides object-placement functions for primitives and asset-based objects, while the rendering host constructs the selected environment and its supporting geometry. Candidate programs are checked for executable structure, object placement, and spatial-constraint satisfaction before rendering. A selected valid program is executed in Blender without adjusting the object layout after geometric verification. Question validation and label computation are performed by the fixed Geometric Oracle, as detailed in~\Cref{app:oracle}.

\paragraph{Multiview rendering.}
We use Blender Cycles with 64 samples per pixel to render four RGB images at $512\times512$ resolution for each accepted scene. The camera system places viewpoints around the scene with approximately quarter-turn angular coverage and adjusts the positions and framing according to scene geometry and object visibility. Indoor and outdoor environments use appropriate background geometry, materials, and lighting. All four observations depict the same realized 3D scene, and no ground-truth answer label is embedded in the images.

\paragraph{Geometric metadata.}
The rendering pipeline records object identities, world-coordinate bounding boxes, positions, sizes, camera poses, and per-view visibility information. Object geometry and camera transforms support world-frame and viewpoint-dependent spatial relations, respectively. Visibility estimates are obtained from projected object geometry and ray-based occlusion checks. The Oracle uses these metadata to evaluate required constraints, question answerability, and the unique answer option. The Solver receives only the rendered views and the validated question, without access to the scene metadata or geometry-derived label.

\subsection{Computational Resources and Training Efficiency}\label{app:compute}
\paragraph{Computational setup.}
Training uses eight NVIDIA A800 GPUs. The Proposer is optimized with four distributed training processes, while the Coder and Solver each use eight. During experience generation, frozen policies perform inference and Blender renders the multiview scenes. The three roles are updated sequentially under the 5/5/15 optimization schedule described in~\Cref{app:training}.

\paragraph{Training time.}
\Cref{tab:app_runtime} compares the training costs of Qwen3.5-4B, Qwen3.5-9B, and InternVL3.5-8B. Data Gen includes preparation of Coder training proposals and generation of Oracle-verified scenes for the Solver. GRPO includes the optimization of all three roles. The single-iteration columns give the phase-level breakdown for one iteration, while the final column reports the accumulated cost of three self-evolution iterations. For the Qwen backbones, the 5/5/15 GRPO durations are obtained by normalizing the logged role-specific training durations by their optimization-step counts; the InternVL3.5-8B entries follow its model-specific cost summary. Totals are reported to two decimal places.

\subsection{Benchmark Protocol and Aggregation}\label{app:eval}
We evaluate the six spatial reasoning benchmarks reported in the main paper. The evaluated task categories and the benchmark-level aggregation rules are summarized in~\Cref{tab:eval_protocol}. All benchmark categories are covered unless a category restriction is explicitly specified.

\paragraph{Input processing and answer evaluation.}
For VSI-Bench, 32 frames are uniformly sampled from each video while preserving their temporal order. Multiple-choice tasks are evaluated using the selected answer option; numerical VSI-Bench tasks use mean relative accuracy (MRA) according to the benchmark's numerical scoring protocol. The remaining image-based benchmarks retain the input image order and their corresponding question and answer formats. Evaluation uses greedy decoding, with answer extraction adapted to the required output type. The same benchmark coverage and scoring rules are used for the compared models.

\paragraph{Score aggregation.}
The six-benchmark average is defined as
\begin{equation}
\mathrm{Avg.}=\frac{1}{6}\sum_{b=1}^{6}\mathrm{Score}(b).
\label{eq:app_benchmark_avg}
\end{equation}
For ViewSpatial-Bench, $\mathrm{Score}(b)$ is the equal-weight mean of Cam-Rel, Per-Rel, and Per-Sce-Sim. For OmniSpatial-test, we use the 561 questions in the three specified Perspective Taking categories and compute
\begin{equation}
\mathrm{Score}_{\mathrm{OmniSpatial}}
=100\,\frac{n_{\mathrm{correct}}}{102+376+83}
=100\,\frac{n_{\mathrm{correct}}}{561}.
\label{eq:app_omni_micro}
\end{equation}
This is a question-level micro average rather than an equal-weight average of the three category accuracies. The six benchmarks are used for evaluation and are not incorporated as external spatial training examples. Main-paper Table~1 reports results averaged over three independent runs with different random seeds.

\subsection{Baseline Reproduction}\label{app:baseline}
We compare Spatial-Zero with SpatialSkill and MM-Zero under the Qwen3.5-4B backbone, using the same six-benchmark evaluation protocol and aggregation rules. As specified in the main paper, the competing methods retain their original configurations except for the backbone. SpatialSkill follows its skill-evolution and inference procedure without parameter updates to the executor, whereas MM-Zero follows its multi-role self-evolution procedure. Their method-specific training or skill-evolution budgets are preserved rather than being redefined in terms of Spatial-Zero's cumulative Solver optimization steps.

%%%%%------------------------------Appendix D
\section{Additional Ablation Details}
\label{app:ablation}
\mainsec{3.3}

\subsection{Independent Human Evaluation of Visual Answerability}
\label{app:human}

\begin{wraptable}{r}{0.48\textwidth}
    \centering
    \vspace{-8pt}
    \small
    \setlength{\tabcolsep}{3pt}
    \caption{Human evaluation of visual answerability
    and Human--Oracle agreement.}
    \label{tab:human_visual_evaluation}
    \begin{tabular}{@{}lcc@{}}
        \toprule
        Metric & Count & Rate \\
        \midrule
        Visually answerable & 196/200 & $98.0\%$ \\
        Human--Oracle agreement & 195/196 & $99.5\%$ \\
        \bottomrule
    \end{tabular}
    \vspace{-8pt}
\end{wraptable}

\paragraph{Protocol.}
Two human annotators independently evaluate 200 questions sampled from those accepted by the full Geometric Oracle. Annotators see only rendered multiview images and four-choice questions, without access to 3D metadata or geometry-derived labels. They first determine whether each question is visually answerable, defined as whether the provided images contain sufficient evidence to identify a unique answer. For visually answerable questions, annotators further select one of the four answer options. The final judgments are confirmed by both annotators, and the selected answers are compared against the geometric ground truth.

\paragraph{Metrics and results.}
Visual Answerability Rate is the proportion of sampled questions judged visually answerable. Human--Oracle Agreement is computed over visually answerable questions. As reported in main-paper \S3.3.3, 196 of the 200 questions are visually answerable ($98.0\%$), and 195 of the 196 human answers agree with the Geometric Oracle ($99.5\%$). The results are summarized in~\Cref{tab:human_visual_evaluation}.

\paragraph{Inter-annotator agreement.}
We further assess inter-annotator consistency using Cohen's $\kappa$ on the independent combined judgments. Each annotation is represented by one of five categories: unanswerable or one of the four answer choices (A--D). The resulting Cohen's $\kappa$ is $0.85$.

% \newpage
%%%%%------------------------------Appendix E
%%%%%------------------------------Appendix E
\section{Additional Self-Evolution Analysis}
\label{app:evolution}
\mainsec{3.4}

\subsection{Iteration Scaling Across Model Sizes}
\label{app:iteration_scaling}

\noindent
\begin{minipage}[t]{0.53\textwidth}
    \vspace{0pt}
    We examine extended self-evolution with Qwen3.5-0.8B,
    Qwen3.5-2B, and Qwen3.5-4B.
    As shown in~\Cref{fig:iteration_scaling}, the three
    models achieve their highest average accuracy across
    six spatial reasoning benchmarks at cumulative Solver
    optimization Steps 15, 30, and 45, respectively.
    Although performance slightly declines after the
    respective peaks, all models remain above their
    base-model performance at Step 60, suggesting that
    the optimal training duration varies with model scale.
\end{minipage}
\hfill
\begin{minipage}[t]{0.43\textwidth}
    \vspace{-12pt}
    \centering
    \includegraphics[
        width=\linewidth,
        trim={0 0 0 5mm},
        clip
    ]{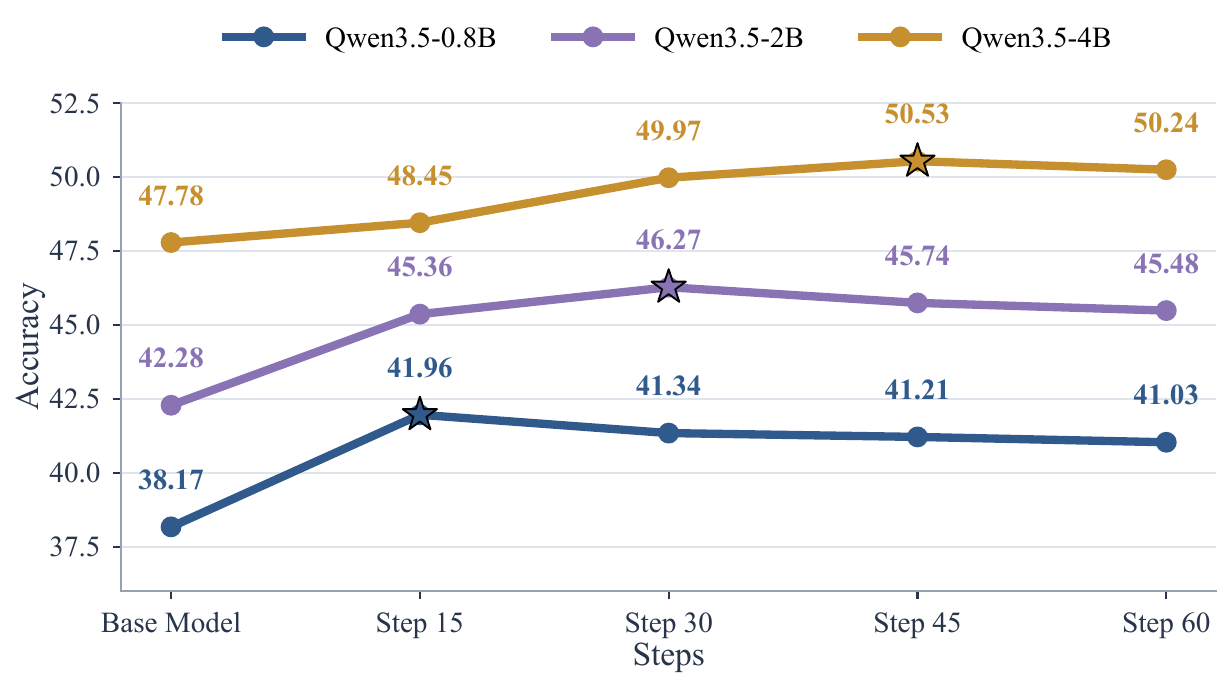}
    \captionof{figure}{Iteration scaling across Qwen3.5 model sizes.}
    \label{fig:iteration_scaling}
\end{minipage}

\par